\pdfoutput=1

\documentclass[11pt]{article}

\usepackage{acl}

\usepackage{times}
\usepackage{latexsym}
\usepackage{graphicx}
\usepackage{booktabs}
\usepackage{subcaption}
\usepackage[ruled,vlined]{algorithm2e}
\usepackage{multirow}
\usepackage{tabularx}
\usepackage{booktabs}

\usepackage{amsmath}
\usepackage{graphicx}
\usepackage{bbm}
\usepackage{subcaption}
\usepackage{tabularx,ragged2e}
\usepackage{multicol,multirow}
\usepackage{enumitem}
\usepackage{soul}
\usepackage[T1]{fontenc}

\usepackage[utf8]{inputenc}

\usepackage{microtype}

\title{Let the Model Decide its Curriculum for Multitask Learning}
\author{Neeraj Varshney,~~ 
  Swaroop Mishra,~~ 
  Chitta Baral
  \\
  Arizona State University \\
  \texttt{\{nvarshn2, srmishr1, cbaral\}}@asu.edu
  }

\begin{document}
\maketitle
\begin{abstract}
Curriculum learning strategies in prior multi-task learning approaches arrange datasets in a difficulty hierarchy either based on human perception or by exhaustively searching the optimal arrangement.
However, human perception of difficulty may not always correlate well with machine interpretation leading to poor performance and exhaustive search is computationally expensive.
Addressing these concerns, we propose two classes of techniques to arrange training instances into a learning curriculum based on difficulty scores computed via model-based approaches. 
The two classes i.e Dataset-level and Instance-level differ in granularity of arrangement.
Through comprehensive experiments with $12$ datasets, we show that instance-level and dataset-level techniques result in strong representations as they  lead to an average performance improvement of $4.17\%$ and $3.15\%$ over their respective baselines.
Furthermore, we find that most of this improvement comes from correctly answering the difficult instances, implying a greater efficacy of our techniques on difficult tasks.
\end{abstract}

\section{Introduction}
\label{introduction_label}
In recent times, Multi-Task Learning (MTL) \cite{caruana1997multitask} i.e. developing one \textit{Generalist} model capable of handling multiple tasks has received significant attention from the NLP community \cite{aghajanyan-etal-2021-muppet,Lu_2020_CVPR, sanh2019hierarchical,clark-etal-2019-bam, mishra-etal-2022-numglue}.
Developing a single model in MTL has several advantages over multiple \textit{Specialist} models as it 
(i) can leverage knowledge gained while learning other tasks and perform better in limited-data scenarios \cite{crammer2012learning, Ruder2017SluiceNL},
(ii) prevents overfitting to a single task, thus providing a regularization effect and increasing robustness
\cite{clark-etal-2019-bam,evgeniou2004regularized}, and 
(iii) provides storage and efficiency benefits because only one model needs to be maintained for all the tasks \cite{bingel-sogaard-2017-identifying}.

Prior work has shown that presenting training instances ordered by difficulty level benefits not only humans but also machines \cite{elman1993learning, xu2020curriculum}. 
Arranging instances in a difficulty hierarchy i.e Curriculum Learning (easy to hard) and Anti-Curriculum Learning (hard to easy) has been studied in MTL setup \cite{mccann2018natural,pentina2015curriculum}.
These techniques arrange datasets either based on human perception of difficulty or by exhaustively searching the optimal arrangement.
However, both these approaches have several limitations. 
Firstly, human perception of difficulty may not always correlate well with machine interpretation, for instance, a dataset that is easy for humans could be difficult for machines to learn or vice-versa. 
Secondly, exhaustive search is computationally expensive and becomes intractable as the number and size of datasets increase.

In this work, we propose two classes of techniques that enable models to form their own learning curriculum in a difficulty hierarchy. 
The two classes i.e Dataset-level and Instance-level differ in the granularity of arrangement.
In dataset-level techniques, we arrange \textbf{datasets} based on the average difficulty score of their instances and train the model sequentially such that all the instances of a dataset are learned together.
In instance-level techniques, we relax the dataset boundaries and order \textbf{instances} solely based on their difficulty scores. 
We leverage two model-based approaches to compute the difficulty scores (Section \ref{scoringTechniques}).

We experiment with 12 datasets covering various NLP tasks and show that instance and dataset-level techniques result in stronger representations with an average performance gain of $4.17\%$ and $3.15\%$ over their respective baselines. 
Furthermore, we analyze model predictions and find that difficult instances contribute most to this improvement implying greater effectiveness of our techniques on difficult tasks.
We note that our techniques are generic and can be employed in any MTL setup.

In summary, our contributions are as follows:
\begin{enumerate}[noitemsep,nosep,leftmargin=*]
    \item \textbf{Incorporating Machine Interpretation of Difficulty in MTL}: We introduce a novel framework for MTL that goes beyond human intuition of sample difficulty and provides model the flexibility to form its own curriculum at two granularities: instance-level and dataset-level.
    
    \item \textbf{Performance Improvement}: We experiment with 12 NLP datasets and show that instance and dataset-level techniques lead to a considerable performance improvement of 4.17\% and 3.15\%. We note that our curriculum arrangement techniques can be used in conjunction with other multi-task learning methods such as dynamic sampling \cite{gottumukkala2020dynamic} and pre-finetuning \cite{aghajanyan-etal-2021-muppet} to further improve their performance.
    
    \item \textbf{Efficacy on Difficult Instances}: Our experiments in low-data regime reveal that the proposed techniques are most effective on difficult instances. This makes them more applicable for real-world tasks as they are often more difficult than abstract toy tasks and provide limited training instances. 
\end{enumerate}

\section{Difficulty Score Computation}
\label{scoringTechniques}
In this section, we describe two model-based difficulty computation methods based on recent works.
\subsection{Cross Review Method}
\label{cross_review}
\citet{xu2020curriculum} proposed a method that requires splitting the training dataset $D$ into $N$ equal meta-datasets ($M_1$ to $M_N$) and training a separate model on each meta-dataset with identical architecture.
Then, each training instance is inferred using the models trained on other meta-datasets and the average prediction confidence is subtracted from $1$ to get the difficulty score. Mathematically, score of instance $i$ ($\in$ $M_k$) is calculated as,
\[
    s_i = 1 - \frac{\sum_{j \in (1,...,N), j \ne k} c_{ji}}{N-1}
\]
where $c_{ji}$ is prediction confidence on instance $i$ given by the model trained on $M_j$. 

\subsection{Average Confidence Across Epochs}
\label{average_confidence}
In this method, the difficulty score is computed by simply averaging the prediction confidences across epochs of a single model and subtracting it from $1$.
\[
     s_i = 1 - \frac{ \sum_{j=1}^{E} c_{ji}}{E}
\]
where the model is trained till $E$ epochs and $c_{ji}$ is prediction confidence of the correct answer given by the model at $j^{th}$ checkpoint.
This method is based on recent works that analyze the behavior of model during training i.e ``training dynamics'' \cite{swayamdipta-etal-2020-dataset} and during evaluation \cite{varshney-etal-2022-ildae}.



\newcommand{\pluseq}{\mathrel{+}=}
\begin{algorithm}
\SetAlgoLined
\textbf{Input:} \\
$D$: the training dataset, \\ 
\{$S_1,...,S_K$\}: splits created from $D$\\
$frac$: fraction of previous split\\
\textbf{Initialization:} Model $M$\\
\For{$i\gets1$ \KwTo $K$}
{
    $train\_data$ = $S_i$ \\
    \For{$j\gets 1$ \KwTo $i-1$}
    {
        $sampled\_S_j$ = Sampler$(S_j, frac)$ \\
        $train\_data \pluseq sampled\_S_j$ \\
    }
    Train $M$ with $train\_data$ \\
}
Train $M$ with $D$
 \caption{General Training Structure}
 \label{general_algo}
\end{algorithm}
\vspace{-0.5cm}
\section{Proposed Techniques}
Addressing the limitations of current approaches highlighted in Section \ref{introduction_label}, we propose two classes of techniques to arrange training instances that 
allow models to form the learning curriculum based on their own difficulty interpretation. 
The technique classes i.e Dataset-Level and Instance-Level leverage difficulty scores computed using methods described in section \ref{scoringTechniques} and follow the general training structure shown in Algorithm \ref{general_algo}. 
The training dataset $D$ is divided into $K$ splits ($S_1,...,S_K$) based on the difficulty score, and model $M$ is trained sequentially on these ordered splits. 
Furthermore, while training the model on split $S_i$, a fraction ($frac$) of instances from previous splits ($S_j (j<i)$) is also included in training to avoid catastrophic forgetting \cite{carpenter1988art} i.e forgetting the previous splits while learning a new split.
Note that $D$ is a collection of multiple datasets in the MTL setup.
The final step requires training on the entire dataset $D$ as the evaluation sets often contain instances of all tasks and difficulty levels.
Dataset-level and Instance level techniques vary in the way splits ($S_1,...,S_K$) are created as described below:

\paragraph{Dataset-level techniques:} In this technique class, each \textbf{dataset} represents a split and is arranged based on the average difficulty score of its instances i.e score of a dataset $D_k$ is calculated as:
\[
    d_k = \frac{\sum_{i \in D_k} s_{i}}{|D_k|}
\]
where, $s_i$ is the difficulty score of instance $i \in D_k$.

\begin{figure}[t]
    \includegraphics[width=7.8cm,height=5cm]{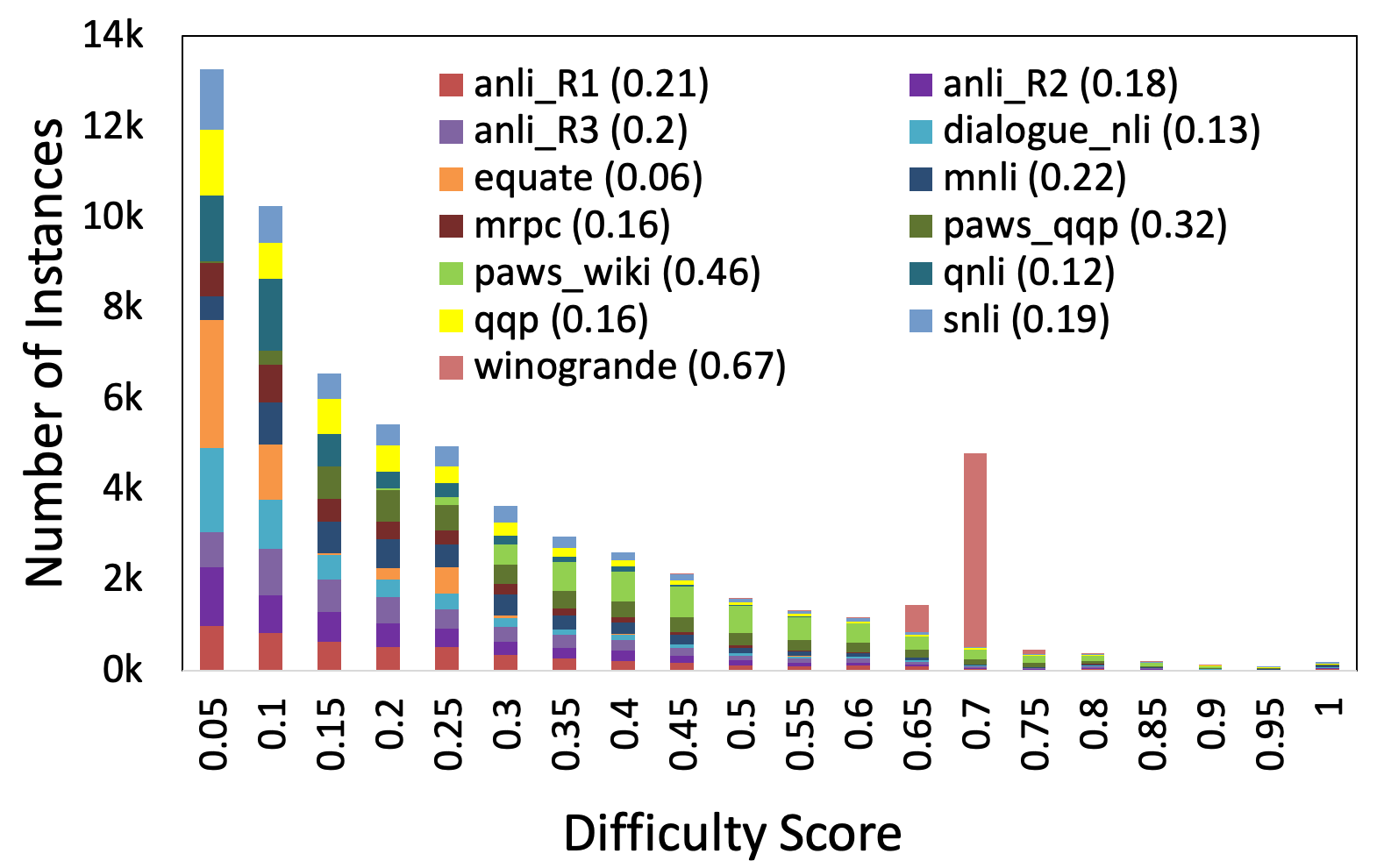}
    \caption{Distribution of instances based on difficulty score computed using Average Confidence method. Difficulty score of datasets are shown in the legends.}
    \label{distribution_of_scores}
    \vspace{-5mm}
\end{figure}

\paragraph{Instance-level techniques:} Here, we relax the dataset boundaries and arrange \textbf{instances} solely based on their difficulty scores.
We study two approaches of dividing instances into splits ($S_1,...,S_K$): \textbf{Uniform and Distribution-based splitting}.
In the former, we create $K$ uniform splits from $D$, while 
in the latter, we divide based on the distribution of scores such that instances with similar scores are grouped in the same split\footnote{\label{refer_supplementary}Refer to Appendix for details}.
The latter approach can result in unequal split sizes as we show in Figure \ref{distribution_of_scores} that the number of instances varies greatly across difficulty scores.

\begin{table*}[]
\resizebox{\textwidth}{!}{%
\begin{tabular}{lll|llllllll|llll}
\toprule
\multicolumn{1}{c}{\multirow{3}{*}{\textbf{Datasets}}} &
\multicolumn{2}{c}{\textbf{Single-Task}} &
  \multicolumn{8}{c}{\textbf{Instance-Level}} &
  \multicolumn{4}{c}{\textbf{Dataset-Level}} \\
\multicolumn{1}{c}{} &
  \multicolumn{2}{l}{} &
  \multicolumn{2}{l}{\underline{Heterogeneous(B)}} &
  \multicolumn{2}{l}{\underline{Uniform}} &
  \multicolumn{2}{l}{\underline{Distribution (D)}} &
  \multicolumn{2}{l}{\underline{D with $frac$=0.4}} &
  \multicolumn{2}{l}{\underline{Random Order(B)}} &
  \multicolumn{2}{l}{\underline{Proposed Order}} \\
\multicolumn{1}{c}{} & EM & F1 & EM & F1 & EM & F1 & EM & F1 & EM & F1 & EM & F1 & EM & F1 \\

\midrule
SNLI & 77.26 & 77.42 & 74.55 & 74.62 & 77.79 & \textbf{77.79} & 77.64 & 77.7 & 77.65 & 77.65 & 77.7 & 77.75 & 78.94 & \textbf{79.05}\\
MNLI Mismatched & 65.98 & 66.12 & 62.07 & 62.14 & 66.14 & 66.3 & 66.71 & \textbf{66.78} & 66.6 & 66.66 & 66.29 & 66.4 & 69.15 & \textbf{69.28}\\
MNLI Matched & 65.33 & 65.45 & 61.23 & 61.36 & 65.85 & 65.96 & 66.91 & \textbf{67.01} & 66.82 & 66.85 & 65.96 & 66.09 & 69.18 & \textbf{69.33}\\
Winogrande & 50 & 50 & 47.34 & 50 & 50.24 & \textbf{50.27} & 50 & 50.12 & 49.82 & 49.85 & 47.99 & 49.85 & 48.37 & \textbf{50.3}\\
QNLI & 74.21 & 74.23 & 66.78 & 66.81 & 70.42 & 70.44 & 71.81 & \textbf{71.81} & 71.38 & 71.38 & 70.35 & 70.39 & 73.75 & \textbf{73.79}\\
EQUATE & 98.99 & 98.99 & 98.99 & 98.99 & 99.14 & 99.21 & 99.57 & \textbf{99.57} & 99.28 & 99.28 & 99.57 & 99.57 & 99.57 & \textbf{99.57}\\
QQP & 80.04 & 80.06 & 75.34 & 75.35 & 78.89 & 78.9 & 79.23 & \textbf{79.25} & 79.11 & 79.12 & 79.23 & 79.26 & 80.27 & \textbf{80.29}\\
MRPC & 80.98 & 80.98 & 74.42 & 74.45 & 74.05 & 74.05 & 75.95 & \textbf{75.98} & 75.4 & 75.4 & 75.73 & 75.77 & 79.08 & \textbf{79.08}\\
PAWS Wiki & 52.45 & 52.49 & 55.92 & 56.01 & 53.15 & 53.16 & 54.39 & 54.47 & 70.59 & \textbf{70.62} & 56.44 & 56.51 & 80.33 & \textbf{80.34}\\
PAWS QQP & 68.25 & 68.41 & 73.03 & 73.03 & 69 & 69 & 71.83 & 71.83 & 78.84 & \textbf{78.84} & 73.08 & 73.12 & 83.46 & \textbf{83.46}\\
ANLI R1 & 42.2 & 42.57 & 38.1 & 38.28 & 42.1 & 42.13 & 45.7 & \textbf{45.7} & 43.2 & 43.33 & 42.9 & \textbf{43.04} & 42.3 & 42.58\\
ANLI R2 & 38.1 & 38.78 & 35 & 35 & 39.8 & \textbf{39.9} & 38.9 & 39.05 & 37.2 & 37.25 & 38.4 & \textbf{38.5} & 36.8 & 36.97\\
ANLI R3 & 39.25 & 39.38 & 36.17 & 36.24 & 38.5 & \textbf{38.62} & 38.17 & 38.24 & 36.5 & 36.56 & 37.92 & \textbf{38.03} & 37.25 & 37.4\\
DNLI & 84.68 & 84.83 & 80.36 & 80.48 & 83.51 & \textbf{83.57} & 83.15 & 83.2 & 82.09 & 82.12 & 82.52 & 82.59 & 82.67 & \textbf{82.73}\\
HANS & - & - & 49.06 & 49.07 & 48.95 & 49.01 & 48.3 & 48.38 & 49.39 & \textbf{49.45} & 48.22 & \textbf{48.27} & 48 & 48.09\\
Stress Test & - & - & 55.28 & 55.44 & 56.2 & 56.31 & 58.66 & \textbf{58.77} & 57.7 & 57.75 & 56.74 & 56.84 & 59.94 & \textbf{60.15}\\

\bottomrule
\end{tabular}
}
\caption{Results on performing curriculum learning using the proposed techniques with difficulty scores computed via Average Confidence approach. Note that $frac$ is 0 unless otherwise mentioned. B means baseline and D with $frac$=0.4 column represents Distribution based splitting with value of $frac$ as 0.4.}
\vspace{-4mm}
\label{results_table}

\end{table*}

\section{Experiments}
\textbf{Datasets:} We experiment with 12 datasets covering various sentence pair tasks, namely, Natural Language Inference (SNLI \cite{bowman-etal-2015-large}, MultiNLI \cite{williams-etal-2018-broad}, Adversarial NLI \cite{nie-etal-2020-adversarial}), Paraphrase Identification (QQP \cite{iyer2017first}, MRPC \cite{dolan2005automatically}, PAWS \cite{zhang-etal-2019-paws}), Commonsense Reasoning (Winogrande \cite{sakaguchi2020winogrande}), Question Answering NLI (QNLI \cite{wang-etal-2018-glue}), Dialogue NLI (DNLI \cite{welleck-etal-2019-dialogue}), and Numerical Reasoning (Stress Test of Equate \cite{ravichander-etal-2019-equate}). 
For evaluation on robustness and generalization parameters, we use HANS \cite{mccoy-etal-2019-right} and Stress Test \cite{naik-etal-2018-stress} datasets.

\paragraph{Setup:} 
We experiment in a low-resource regime limiting the number of training instances of each dataset to $5000$. This enables evaluating our techniques in a fair and comprehensive manner as transformer models achieve very high accuracy when given large datasets.
Furthermore, inspired by decaNLP \cite{mccann2018natural}, we reformulate all the tasks in our MTL setup as span identification Question Answering tasks\footnotemark[1]. 
This allows creating a single model to solve the tasks that originally have different output spaces. 
\paragraph{Implementation Details:}
We use three values of $frac$: 0, 0.2, and 0.4 (refer Algorithm 1), $N=5$ (in Cross Review method), and $E=5$ (in Average Confidence method). 
For distribution-based splitting, we experiment by dividing $D$ into 3 and 5 splits\footnotemark[1]. 
These hyper-parameters are selected based on development dataset performance.
\paragraph{Baseline Methods:}
In MTL, \textit{heterogeneous} batching where all the datasets are combined and a batch is randomly sampled
has been shown to be much more effective than \textit{homogeneous} and \textit{partitioned} batching strategies \cite{gottumukkala2020dynamic}. 
Therefore, we use it as the baseline for instance-level techniques.
For dataset-level techniques, we generate multiple dataset orders and take the average performance as the baseline. 
We average these baseline scores across 3 different runs. 

\section{Results and Analysis}
\label{results_section}
\begin{figure}
    \centering
    \includegraphics[width=5cm,height=4cm]{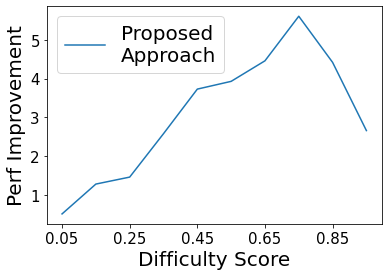}
    \caption{Performance improvement vs Difficulty score for dataset level techniques.}
    \label{fig:analysis}
\end{figure}

Table \ref{results_table} shows the 
efficacy of our proposed curriculum learning techniques.
\paragraph{Performance Improvement: } \textit{Instance and Dataset-level techniques achieve an average improvement of 4.17\% and 3.15\% over their respective baseline methods.}
This improvement in consistent across all the datasets and also outperforms single-task performance in most cases.
Furthermore, we find that \textit{models leveraging Average Confidence method (\ref{average_confidence}) outperform their counterparts using the Cross Review method (\ref{cross_review})\footnotemark[1] rendering Average Confidence approach as more effective both in terms of performance and computation as Cross Review requires training multiple models (one for each meta-dataset)}.
\paragraph{Uniform Vs Distribution based splitting: } \textit{In instance-level experiments, distribution-based splitting shows slight improvement over uniform splitting. }
We attribute this to the superior inductive bias resulting from the collation of instances with similar difficulty scores to the same split.
\paragraph{Effect of adding instances from previous splits:}
\textit{For dataset-level techniques, we find that it does not provide any improvement}. This is because all the instances of a dataset are grouped in a single split therefore, adding instances from other splits doesn't contribute much to the inductive bias.
Furthermore, \textit{in the case of instance-level, it leads to a performance improvement} because previous splits contain instances of the same dataset hence, providing the inductive bias.
\paragraph{Difficulty Scores Analysis:} 
Figure \ref{distribution_of_scores} shows the distribution of training instances of all datasets with difficulty scores computed using Average confidence (\ref{average_confidence}) method. 
This distribution reveals that instances across datasets and within every dataset vary greatly in difficulty as they are widely spread across the difficulty scores. 
Comparing the average difficulty score of all datasets (shown in legends of Figure \ref{distribution_of_scores}) shows that \textit{Equate and QNLI are easy-to-learn while PAWS and Winogrande are relatively difficult-to-learn.}
Furthermore, \textit{around 32\% of the training instances get assigned a difficulty score of $\leq 0.1$ hinting at either the presence of dataset artifacts or the inherent easiness of these instances.} 
A similar observation is made with Cross Review method with the percentage being 37\%.
\paragraph{Test Set Analysis:} We also compute difficulty scores of test instances and plot the performance improvement achieved by our approach over the baseline method for every difficulty score bucket in Figure \ref{fig:analysis}.
We find that \textit{the proposed technique is effective especially on instances with high difficulty scores}. 
This implies a greater efficacy of our techniques on tasks that contain difficult instances.


\section{Conclusion}
In this paper, we proposed two classes of techniques for MTL that allow models to form the learning curriculum based on their own interpretation of difficulty. 
Comprehensive experiments with $12$ datasets showed that our techniques lead to a performance improvement of $4.17\%$ and $3.15\%$. Furthermore, we found that difficult instances contribute most to this improvement, implying a greater efficacy of our techniques on difficult tasks. 
We hope that our techniques and findings will foster development of stronger multi-task learning models as our curriculum arrangement techniques can be used in conjunction with other multi-task learning methods such as dynamic sampling \cite{gottumukkala2020dynamic} and pre-finetuning \cite{aghajanyan-etal-2021-muppet} to further improve their performance.

\section*{Acknowledgements}
We thank the anonymous reviewers for their insightful feedback. This research was supported by DARPA SAIL-ON program.

\bibliography{anthology,custom}
\bibliographystyle{acl_natbib}
\appendix
\section*{Appendix}

\section{Statistics of Evaluation Datasets}
In this work, we experiment with 12 datasets spanning over several NLU tasks. Table \ref{tab:datasets} shows the statistics of the evaluation sets.

\begin{table*}
    \centering
    \begin{tabular}{p{13cm}p{2.1cm}}
    \toprule
        \textbf{Context -- Question} &
        \textbf{Datasets}
        \\

    \midrule
    \textbf{C: }Kyle doesn't wear leg warmers to bed, while Logan almost always does. he is more likely to live in a colder climate. \textbf{false}, or true ? \\
    \textbf{Q: }Kyle is more likely to live in a colder climate.
    & Winogrande \\
    
    \midrule
    \textbf{C: }In order for an elevator to be legal to carry passengers in some jurisdictions it must have a solid inner door. \textbf{false}, or true ? \\
    \textbf{Q: }What is another name for a freight elevator?  Does the context sentence contain answer to this question ?
    & QNLI \\
    
    \midrule
    \textbf{C: }What makes a great problem solver? false, or \textbf{true}? & QQP, MRPC, \\
    \textbf{Q: }How can I be a fast problem solver?  Are the two sentences semantically equivalent?
    & PAWS \\
    
    \midrule
    \textbf{C: }i sell miscellaneous stuff in local fairs . \textbf{contradiction}, or neutral, or entailment ? \\
    \textbf{Q: }i used to work a 9 5 job as a telemarketer . Consistency of the dialogues ?
    & DNLI \\
    
    \midrule
    \textbf{C: }205 total Tajima' s are currently owned by the dealership. contradiction, or neutral, \textbf{entailment} ? \\
    \textbf{Q: }less than 305 total Tajima' s are currently owned by the dealership.
    & Equate \\
    
    \midrule
    \textbf{C: }Two collies are barking as they play on the edge of the ocean contradiction, or neutral, or \textbf{entailment} ? & SNLI, MNLI, ANLI \\
    \textbf{Q: }Two dogs are playing together.
    &  \\

    \bottomrule

    \end{tabular}
    \caption{Illustrative examples (Context (C) - Question (Q) pairs) of different types of training datasets considered in this work. We transform all these datasets to Question-Answering format in order to use a single model for all these tasks. Answers are highlighted in \textbf{bold}.}
    \label{tab:example_table}
\end{table*}

\begin{figure*}[t]
\centering
    \begin{subfigure}{0.48\textwidth}
         \includegraphics[width=7.8cm,height=5cm]{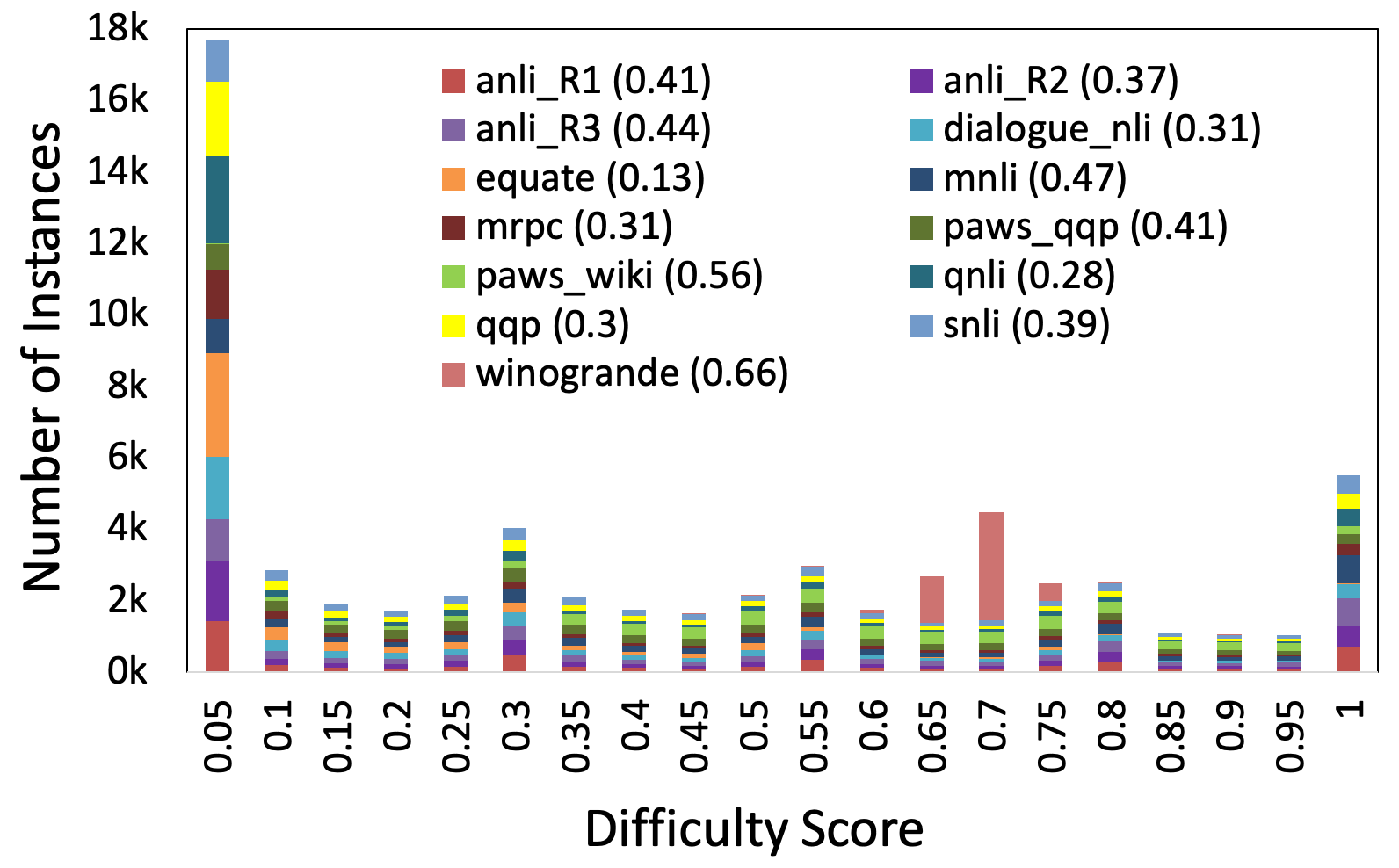}
         \caption{Cross Review approach}
    \end{subfigure}
    \begin{subfigure}{0.48\textwidth}
        \includegraphics[width=7.8cm,height=5cm]{Pictures/difficulty_average_conf.png}
        \caption{Average Confidence approach}
    \end{subfigure}
    \caption{Distribution of instances based on difficulty scores computed using (a) Cross Review approach and (b) Average Confidence approach.}
    \label{appendix_distribution_of_scores}
\end{figure*}

\begin{table*}[]
\begin{tabular}{lll|llllllll|llll}
\toprule
\multicolumn{1}{c}{\multirow{3}{*}{\textbf{Datasets}}} &
  \multicolumn{2}{c}{\textbf{Instance-Level}} &
  \multicolumn{4}{c}{\textbf{Dataset-Level}} \\

\multicolumn{1}{c}{} &
  \multicolumn{2}{l}{\underline{Uniform Splitting + Prev}} &
  \multicolumn{2}{l}{\underline{Proposed Order with $frac$=0.4}} &
  \multicolumn{2}{l}{\underline{AC on Proposed Order}} \\
\multicolumn{1}{c}{} & EM & F1 & EM & F1 & EM & F1  \\

\midrule
SNLI & 76.19 & 76.2 & 77.09 & 77.11 & 77 & 77.02\\
MNLI Mismatched & 64.54 & 64.55 & 65.83 & 65.85 & 65.36 & 65.41\\
MNLI Matched & 63.63 & 63.64 & 66.06 & 66.08 & 64.72 & 64.77\\
Winogrande & 50.48 & 50.48 & 50.6 & 50.94 & 48.43 & 49.79\\
QNLI & 68.16 & 68.17 & 71.24 & 71.25 & 72.23 & 72.26\\
EQUATE & 99.71 & 99.71 & 99.43 & 99.43 & 99.57 & 99.57\\
QQP & 77.61 & 77.61 & 79.32 & 79.32 & 79.68 & 79.71\\
MRPC & 72.15 & 72.15 & 76.07 & 76.07 & 77.55 & 77.55\\
PAWS Wiki & 52.11 & 52.13 & 69.48 & 69.48 & 52.92 & 52.95\\
PAWS QQP & 68.7 & 68.7 & 69.75 & 69.75 & 66.62 & 66.69\\
ANLI R1 & 41.9 & 41.93 & 43.8 & 43.88 & 44.7 & 44.8\\
ANLI R2 & 37.8 & 37.85 & 36.8 & 36.83 & 37.4 & 37.5\\
ANLI R3 & 37.58 & 37.62 & 36.5 & 36.53 & 36.83 & 36.83\\
DNLI & 82.55 & 82.58 & 83.64 & 83.66 & 81.83 & 81.93\\
HANS & 49.76 & 49.77 & 48.24 & 48.28 & 50.25 & 50.26\\
Stress Test & 56.07 & 56.09 & 57.55 & 57.57 & 58.79 & 58.87\\
\bottomrule
Average & 62.43 & 62.45 & 64.46 & 64.5 & 63.37 &	63.49  \\

\bottomrule
\end{tabular}
\caption{Results of instance-level and dataset-level techniques.}
\label{results_table_sup}

\end{table*}

\section{Implementation Details}
We use the huggingface implementation of BERT-Base model, batch size $16$, learning rate $5e-5$ for our experiments.
We use three values of $frac$: 0, 0.2, and 0.4 (refer Algorithm 1), $N=5$ (in Cross Review method), and $E=5$ (in Average Confidence method). 
For distribution based splitting, we experiment by dividing $D$ into 3 and 5 splits. 
The results reported in the paper are for 3 splits.
These hyper-parameters are selected based on performance on the dev dataset.
We adjust the per gpu training batch size and gradient accumulation accordingly to fit in our 4 Nvidia V100 16GB GPUs. We keep the maximum sequence length of 512 for our experiments to ensure that the model uses the full context.

\begin{table}
\small
    \centering
    \begin{tabular}{p{1.7cm}p{1cm}|p{1.8cm}p{0.9cm}}
     \toprule
        \textbf{Dataset} & \textbf{Size} & \textbf{Dataset} & \textbf{Size}\\
         \midrule
        SNLI & 9824 & MNLI & 19645 \\
        Winogrande & 1654 & QNLI & 5650 \\ 
        PAWS qqp & 671 & PAWS wiki & 7987 \\ 
        MRPC & 1630 & ANLI R1 & 1000 \\ 
        ANLI R2 & 1000 & ANLI R3 & 1000 \\ 
        DNLI & 16408 & HANS & 30000 \\ 
        Equate & 696 & QQP & 40371 \\
        Stress Test & 136464 &  &  \\
        
        \bottomrule
    \end{tabular}
    \caption{Statistics of the evaluation datasets.}
    \label{tab:datasets}
\end{table}

\begin{table}
    \centering
    \begin{tabular}{p{1.5cm}p{1.5cm}|p{1.25cm}p{1.25cm}}
     \toprule
        \textbf{Difficulty Score} & \textbf{Instances} & \textbf{Random Order} & \textbf{Proposed Order} \\
         \midrule
        0.1 & 63736 & 94.86 & 93.77\\
        0.2 & 18703 & 87.8 & 85.55\\
        0.3 & 28035 & 81.85 & 79.85\\
        0.4 & 17238 & 74.5 & 72.81\\
        0.5 & 21502 & 65.03 & 65.84\\
        0.6 & 17338 & 57.69 & 57.94\\
        0.7 & 21255 & 46.75 & 48.92\\
        0.8 & 18058 & 38.36 & 44.05\\
        0.9 & 22327 & 26.8 & 33.07\\
        1 & 46008 & 9.17 & 14.05\\
                
        \bottomrule
    \end{tabular}
    \caption{Comparing performance of random order and the proposed order (developed using our curriculum strategy) across all difficulty scores for instance level techniques. }
    \label{tab:analysis_table}
\end{table}

\section{Dataset Examples}

Table \ref{tab:example_table} shows examples of different types of datasets used in this work. 
We transform all these datasets to Question-Answering format in order to use a single model for all these tasks.

\section{Difficulty Scores}
Figure \ref{appendix_distribution_of_scores} shows the distribution of difficulty scores computed using Cross Review and Average Confidence approach.

\section{Results}
Table \ref{results_table_sup} shows the results of instance-level and dataset-level techniques.

\section{Analysis}

In table \ref{tab:analysis_table}, we compare the performance of random order and the proposed order (developed using our curriculum strategy) across all difficulty scores for instance level techniques.


\section{Scheduling in Multi-task Learning}
Scheduling in multi-task learning has attracted a lot of attention, especially for the machine translation task \cite{zaremoodi-haffari-2019-adaptively, kiperwasser-ballesteros-2018-scheduled, jean2019adaptive}.
Such approaches can be adapted for our tasks and can further improve the multi-task performance. We leave these explorations for future work. 

\section{Limitations of Computing Difficulty Scores using Model-based Techniques}
In addition to arranging the training instances into a learning curriculum, computing difficulty scores using model-based techniques has shown its benefits in several other areas, such as improving selective prediction ability \cite{varshney-etal-2022-towards}, understanding training dynamics \cite{swayamdipta-etal-2020-dataset}, and efficient evaluations \cite{varshney-etal-2022-ildae}.
However, these techniques present a few challenges:

\begin{enumerate}[noitemsep,nosep,leftmargin=*]
    \item \textbf{Computation:} They involve calculating the difficulty scores of training instances which requires additional computation. However, this computation is only required during training and not required during inference. Hence, it does not add any computational overhead at inference time when deployed in an application.
    
    \item \textbf{Noisy Instances: } Training instances that are wrongly annotated/noisy will most certainly get assigned a very high difficulty score and hence will be learned at the end during training. This is unlikely to hamper learning when the number of noisy instances is small. However, it may negatively impact the model's learning when the training dataset has a non-trivial number of noisy instances. We plan to investigate this in our future work.
\end{enumerate}

\end{document}